\documentclass[runningheads]{llncs}
\usepackage{accv}
\usepackage{accvabbrv}
\usepackage{graphicx}
\usepackage{booktabs}
\usepackage[accsupp]{axessibility}
\usepackage{hyperref}
\usepackage{orcidlink}

\newcommand{\cls}[1]{\textsc{#1}}

\begin{document}

\title{Small Data in Long-Term Environmental Vision: Annotation Noise, Gamification, and Label-Budget Scaling}
\titlerunning{Small Data in Long-Term Environmental Vision}

\author{Bryceton Bible\inst{1} \and Hairong Qi\inst{1} \and Masahiro Toyoura\inst{2}}
\authorrunning{B.~Bible et al.}
\institute{University of Tennessee, Knoxville, TN, USA\\
\email{bbible3@vols.utk.edu, hqi@utk.edu} \and
University of Yamanashi, Kofu, Japan\\
\email{mtoyoura@yamanashi.ac.jp}}

\maketitle

\begin{abstract}
Long-running environmental camera networks produce abundant images but few reliable annotations. Using FujiView, a network of webcams around Mount Fuji, we study three linked questions: whether gamification affects annotation participation and agreement, how annotators disagree on ordinal visibility labels, and how label budget affects downstream classification. Across seven nonrandomized sessions with 22 labelers, gamification showed a more promising pattern for sustained participation than for labeling speed or agreement. In the closest remote-session comparison, the session recorded as gamified had longer active labeling time, while throughput and reference agreement remained similar. Annotation disagreement was substantial at fine granularity: crowd votes matched the single-annotator reference on 49.1\% of exact visibility classes, versus 90.2\% after grouping them into \cls{Good}/\cls{Poor}. In five-replicate label-budget experiments, models using 25\% of the available label pool achieved held-out exact accuracy only 2.8 percentage points below the full-pool model on average. Together, these findings highlight the importance of designing reliable annotation targets, evaluating engagement mechanisms without assuming causal benefits, and using empirical scaling curves to plan labeling effort.
\end{abstract}

\section{Introduction}
\label{sec:intro}

Fixed outdoor cameras record the atmosphere between lens and landmark for years, and such networks already exist for tourism, traffic and weather reporting~\cite{jacobs2007amos,babari2012visibility}. Collecting a dataset from them is not the hard part: the FujiView archive studied here grew to more than 420{,}000 frames in under seventeen months without human effort; obtaining consistent visibility labels is. Scenic visibility of Mount Fuji, the quantity FujiView predicts~\cite{bible2026fujiview}, is a human judgment on a four-class ordinal scale (plus \cls{Bad} for unusable frames), and the annotation campaign studied here collected 23{,}894 human labels, one for every eighteen archived frames. Every model trained on it is a small-data model: the images are free, the labels are scarce, expensive and noisy.

We ask three questions a group building a similar dataset must answer: does gamifying the annotation tool change how much people label and how carefully (\cref{sec:system,sec:sessions}); where does disagreement in such labels come from, and which coarser targets survive it (\cref{sec:archive,sec:noise}); and how much downstream accuracy does a given number of labels buy (\cref{sec:utility}). We built a web labeling tool with a shared \emph{progress wall} and leaderboard in the tradition of games with a purpose~\cite{vonahn2004labeling,vonahn2008designing} and gamified crowdsourcing~\cite{deterding2011gamification,mekler2017elements,landers2017leaderboards}, ran seven timed sessions with 22 labelers in February and March 2026 with gamification on and off, in the laboratory and remotely, and served a reference set at a fixed rate so that every session yields an agreement score. Our contributions are an instrumented gamified annotation system and a descriptive comparison of participation, throughput, and agreement across sessions; a characterization of disagreement in ordinal environmental-vision labels; controlled experiments on renewed supervision and label-budget scaling across replicated runs; and practical guidance for planning similar campaigns (\cref{sec:sessions,sec:noise,sec:utility}).

\section{Related Work}
\label{sec:related}

\paragraph{Gamification and collaborative annotation.}
The ESP Game established image labeling as a game with a purpose~\cite{vonahn2004labeling,vonahn2008designing}. Gamification~\cite{deterding2011gamification} has since been applied to microtasks, crowdsourcing, and citizen science~\cite{feyisetan2015microtasks,morschheuser2017gamified,bowser2013gamification,tinati2017eyewire}, with context-dependent effects~\cite{hamari2014does}. Points and leaderboards can increase output without improving quality~\cite{mekler2017elements}, while leaderboard effects have been explained through goal setting~\cite{landers2017leaderboards}. We examine a domain-specific, shared progress wall and distinguish participation duration from per-label throughput and agreement, without treating our observational sessions as causal evidence.

\paragraph{Annotation disagreement and target reliability.}
Crowd labels are useful but noisy~\cite{snow2008cheap,sheng2008label}; aggregation and annotator models can address disagreement~\cite{dawid1979observer,whitehill2009vote,raykar2010crowds}. Errors persist even in crowd-verified benchmarks~\cite{deng2009imagenet,northcutt2021pervasive}, and disagreement may itself carry information~\cite{uma2021disagreement,plank2022variation,peterson2019uncertainty}. We test the reliability of ordinal visibility categories through crowd--reference and same-annotator comparisons, including coarser decision groupings.

\paragraph{Environmental vision and label budgets.}
Webcam, phenology, and camera-trap archives generate imagery faster than experts can label it~\cite{jacobs2007amos,laffont2014transient,richardson2018phenocam,swanson2015serengeti,beery2018terra}; volunteer labels can be assessed against expert judgments~\cite{lintott2008galaxyzoo,kosmala2016quality}. Camera-based visibility has been estimated with measured targets~\cite{graves2014visibility,palvanov2019visnet} or pairwise judgments~\cite{lepetit2020visibility}. Building on data-scaling, transfer, and active-learning research~\cite{sun2017revisiting,hestness2017scaling,kornblith2019transfer,brigato2021small,settles2012active,ren2022survey}, we measure the downstream returns to additional crowd labels using replicated learning curves on a long-term webcam archive.

\section{The FujiView Archive and Its Label Problem}
\label{sec:archive}

\begin{figure}[t]
\centering
\includegraphics[width=0.94\textwidth]{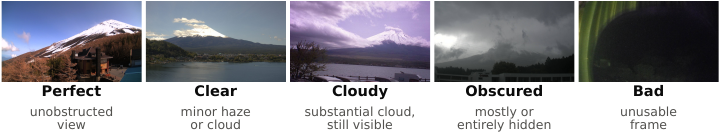}
\caption{The five visibility classes on archive frames (reference label confirmed by at least two agreeing crowd votes; \cls{Bad}: a crowd \cls{Bad} vote and pre-filter exclusion).}
\label{fig:montage}
\end{figure}

The classes (\cref{fig:montage}), taken from the original study~\cite{bible2026fujiview}, are human-facing decision categories rather than estimates of a physical percentage: \cls{Perfect} is an essentially unobstructed view, \cls{Clear} a usable view with minor haze or cloud, \cls{Cloudy} substantial cloud overlap with the mountain still visible, \cls{Obscured} a mountain mostly or entirely hidden, and \cls{Bad} a frame unusable because of darkness or camera failure. The first four cut a continuous phenomenon at boundaries chosen for a traveler who needs to know whether a view is pristine, usable, compromised or unavailable; \cref{sec:noise} suggests that a future interface could collect an anchored 0--100 estimate and derive classes afterward.

FujiView fetches a frame from each public webcam about every half hour~\cite{bible2026fujiview}; in September 2026 the archive held 424{,}853 frames from 34 cameras (since 25 April 2025). By early 2026 a second campaign was needed to renew the original YOLOv8n classifier, trained by the original study on 6{,}149 hand-labeled frames~\cite{bible2026fujiview}, and to build a held-out set sharing no frame with its training data. Of the 257{,}103 frames exposed to annotators, a rule-based pre-filter for near-black frames removed 74{,}359 (28.9\%); even so, at the campaign throughput measured below (about 440 labels per active labeler-hour), labeling the rest once would take roughly 400 labeler-hours. A 2{,}010-frame subset captured on 1--3 May 2025 and labeled that month by one author (the reference annotator) with the four visibility classes (no \cls{Bad}) serves as a \emph{gold} reference, served with probability 0.10 so that every session yields an agreement score; we say \emph{agreement} rather than accuracy because the reference is one author's judgment (\cref{sec:noise}).

\section{A Gamified Annotation System}
\label{sec:system}

\begin{figure}[t]
\centering
\includegraphics[width=0.65\textwidth]{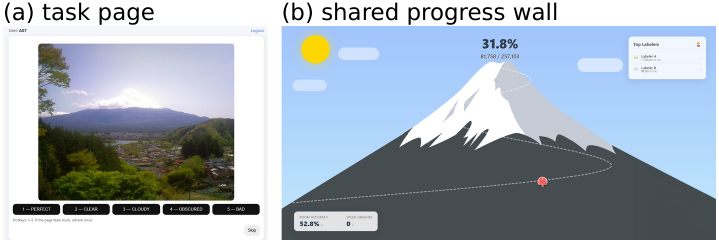}
\caption{Task page and progress wall (rendered from deployed templates with a pseudonymised 3 March 2026 snapshot). (a) Five-class labeling task with hotkeys and skip. (b) Shared mountain-path progress, leaderboard standing, and aggregate pace and reference agreement. The gamified display made collective progress visible without changing the labeling task.}
\label{fig:ui}
\end{figure}

A labeler is shown one frame and five hotkeyed class buttons (\cref{fig:ui}a); a label or a skip requests the next task, and the server times each decision. The selector serves a gold frame with probability 0.10 (two votes per gold frame, up to four after a disagreement) and otherwise an unlabeled frame, one label per frame, so 22{,}884 image records received 23{,}894 labels and 676, all gold, received two or more (gold frames are also regular records). The gamification layer adds nothing to the task page: a shared display in the room, or an accessible browser tab for remote participants, provides the progress wall of \cref{fig:ui}b, which targets competence (visible standing) and relatedness (a shared goal)~\cite{sailer2017motivates}. A separate mobile successor explores gacha-style collectible views as an additional engagement mechanism; this feature is not included in the present evaluation.

\section{Annotation Sessions}
\label{sec:sessions}

\begin{table}[t]
\centering
\caption{The seven sessions (2026). Active min: median per labeler of time inside labeling bouts (gaps $\le5$~min); Labels/h: session total over summed active time; s/label: median decision time (0.5--60~s); Exact, $\pm1$: agreement of gold-frame votes with the reference (exact; within one class) and the number of four-class votes. $^{a}$Recorded as gamified; wall available but individual viewing was not logged. $^{b}$Network failure; excluded from the comparisons.}
\label{tab:sessions}
\scriptsize
\setlength{\tabcolsep}{3pt}
\resizebox{\textwidth}{!}{\begin{tabular}{@{}llcrrrrrrcc@{}}
\toprule
Session & Modality & Gam. & Labelers & Labels & Skips & Active min & Labels/h & s/label & Exact ($n$) & $\pm 1$ \\
\midrule
S1 (Feb 5) & lab & -- & 2 & 1,208 & 6 & 32 & 1130 & 1.8 & 59\% (112) & 96\% \\
S2 (Mar 2) & lab & \checkmark & 4 & 3,895 & 21 & 105 & 582 & 3.3 & 46\% (333) & 95\% \\
S3 (Mar 3) & lab & \checkmark & 8 & 6,409 & 52 & 73 & 631 & 3.0 & 45\% (601) & 95\% \\
S4 (Mar 4--5) & remote & \checkmark$^{a}$ & 8 & 6,387 & 59 & 118 & 397 & 3.9 & 50\% (630) & 96\% \\
S5 (Mar 18) & remote & -- & 7 & 4,388 & 402 & 95 & 415 & 3.7 & 51\% (349) & 95\% \\
S6$^{b}$ (Mar 19) & lab & \checkmark & 7 & 388 & 2 & 55 & 66 & 24.5 & 62\% (39) & 100\% \\
S7 (Mar 20--21) & remote, async. & -- & 1 & 1,219 & 0 & 125 & 584 & 3.0 & 53\% (104) & 96\% \\
\bottomrule
\end{tabular}
}
\end{table}

\paragraph{Design.}
Sessions (\cref{tab:sessions}) used student volunteers paid hourly for active participation (23 accounts: 21 students, one with two logins, and a session operator with 100 labels, mostly in the pilot): a two-person pilot (S1), laboratory sessions with the wall on a shared monitor (S2, S3, S6), a remote session recorded as gamified, with the wall available in a browser tab (S4), and its non-gamified counterpart (S5), and one asynchronous remote labeler without the wall (S7); the pre-filter was enabled from S3. They were not randomised trials: participants, group sizes and times of day differ, and 9 of the 22 labelers took part in more than one session, four under both conditions, so we report descriptively. \emph{Throughput} (labels per active labeler-hour~\cite{vonahn2008designing}), \emph{decision time} and \emph{agreement} are defined in \cref{tab:sessions}; agreement is also reported on the grouping \cls{Good} (\cls{Perfect}, \cls{Clear}) versus \cls{Poor}, excluding the 229 \cls{Bad} votes on gold frames because the reference has no \cls{Bad} class.

\paragraph{Throughput and decision time.}
\Cref{tab:sessions} suggests a more promising pattern for time on task than for labeling pace: gamified sessions were not faster within either modality. Remotely, where the comparison is closest to controlled, the gamified S4 and the non-gamified S5 had similar group sizes, decision times (3.9 versus 3.7~s) and throughput (397 versus 415 labels per active hour); in the laboratory, the gamified S2 and S3 were slower than the two-person pilot (a practised student and the session operator), an upper bound rather than a baseline. What differed more was time on task: median active time per labeler was 32~min in the pilot, 105 and 73~min in the gamified in-person sessions, 118~min in the gamified remote session against 95~min in its non-gamified counterpart, and 125~min for the single asynchronous labeler (S7); the in-person gamified sessions were faster per active hour than S5 but differ from it in modality, so the cleanest contrast is S4 against S5, where the gamified session ran longer at a similar pace (its window was also longer, 4.1 versus 2.9~h). This is consistent with controlled studies in which leaderboards increased the amount of work done~\cite{landers2017leaderboards,mekler2017elements} and with participants' unprompted remarks about wanting to climb the leaderboard or to find particular scenes such as a chosen viewpoint at sunrise, which point to competition and discovery as mechanisms worth testing; neither the design nor these informal remarks establish causation. Skips were rare in every session (0--0.9\% of labels plus skips) except S5 (402, 8.4\%), 354 of them in its first half hour of labeling, which began with repeated server restarts. Thus, participation duration and throughput per active labeler-hour should be assessed separately when evaluating engagement mechanisms.

\paragraph{Agreement.}
The second result is that agreement was similar across conditions. Exact agreement with the reference was 45--50\% in the gamified sessions and 51--53\% in the non-gamified S5 and S7, a difference inside the spread between individual labelers (42--62\%); within-one-class agreement was 95--96\% everywhere and \cls{Good}/\cls{Poor} agreement 89--93\%, so the fine-grained labels did not differ appreciably with gamification or remote participation, consistent with Mekler \etal's finding that game elements change quantity rather than quality~\cite{mekler2017elements}. Pooling all 2{,}168 four-class votes on gold frames gives 49.1\% exact, 95.4\% within one class and 90.2\% \cls{Good}/\cls{Poor} agreement. Serving reference frames throughout the campaign allowed agreement to be monitored across sessions, though it does not establish correctness.

\section{Where the Noise Comes From}
\label{sec:noise}

\begin{figure}[t]
\centering
\begin{minipage}[t]{0.40\textwidth}
\centering
\includegraphics[width=\textwidth]{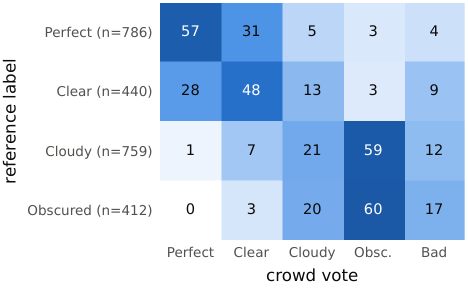}
\end{minipage}\hfill
\begin{minipage}[t]{0.46\textwidth}
\centering
\includegraphics[width=\textwidth]{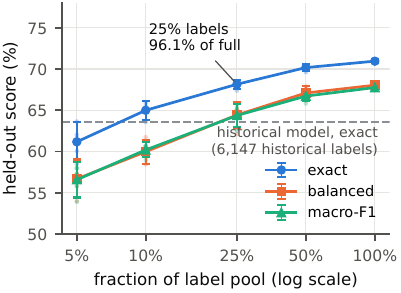}
\end{minipage}
\caption{Disagreement and diminishing returns. Left: row-normalised crowd votes versus the single-annotator reference (2{,}397 gold-frame votes); confusion concentrates between adjacent classes. Right: YOLO11n held-out scores versus label-pool percentage (log-scaled axis): at 25\% of labels, exact accuracy is 2.8 points below the full-pool mean. Means and 95\% intervals over five nested replicates; faint points show individual runs, dashed line the historical model.}
\label{fig:confusion}
\label{fig:budget}
\end{figure}

\Cref{fig:confusion} (left) shows that disagreement is concentrated at two boundaries: reference \cls{Cloudy} frames received the \cls{Obscured} vote 59\% of the time and the \cls{Cloudy} vote only 21\%, and reference \cls{Perfect} frames were voted \cls{Clear} in 31\% of cases. Even with written rules (``substantial cloud overlap'' versus ``mostly or completely hidden''), the boundary between adjacent classes depends on cloud texture, haze and light, which different people resolve differently; the \cls{Cloudy}/\cls{Obscured} confusion appears in every session and for every labeler with more than thirty gold votes (\cls{Cloudy} recall at most 41.2\%). On the 676 frames labeled by two or more people (five-class votes including \cls{Bad}), 58\% were unanimous, pairwise exact agreement was 47\% (58\% on each frame's first two votes), within-one-class agreement 79\%, and Cohen's $\kappa$ between labeler pairs sharing at least thirty frames had median 0.31 (range 0.14--0.46)~\cite{cohen1960kappa,artstein2008agreement}.

\emph{Reference reliability.} Even the reference annotator changed 54\% of exact labels when re-labeling 232 frames in August 2025 that they had labeled in May, while 94\% of the 161 frames given a visibility class both times stayed within one class and 89\% kept their \cls{Good}/\cls{Poor} grouping. The coarse decision, by contrast, is comparatively stable: \cls{Good}/\cls{Poor} agreement with the reference is 89--93\% in every session and 83--95\% (median 91\%) for every labeler with more than thirty gold votes. Exact class identity is therefore not a clean ground-truth target here: target granularity should follow measured agreement, and exact accuracy should be reported alongside within-one-class and \cls{Good}/\cls{Poor} accuracy.

\section{Downstream Utility and Label-Budget Scaling}
\label{sec:utility}

\begin{table}[t]
\centering
\caption{Held-out classification (\%) on 3{,}720 crowd-labeled frames from 33 cameras, disjoint from training. Historical is one checkpoint; A--D are means $\pm$ SD over five seeds. Historical and A--C use YOLOv8n; D uses YOLO11n. Bal.: balanced accuracy; F1: macro-F1; G/P: \cls{Good}/\cls{Poor}; $\pm1$: within-one-class accuracy; P/CL/CY/O: recall for \cls{Perfect}/\cls{Clear}/\cls{Cloudy}/\cls{Obscured}. Rows compare recipe (Historical$\rightarrow$A), renewed labels \emph{and} image samples (A$\rightarrow$B), label count (B$\rightarrow$C), and backbone (C$\rightarrow$D); the backbone swap yields the smallest observed gain.}
\label{tab:downstream}
\scriptsize
\setlength{\tabcolsep}{2.5pt}
\resizebox{\textwidth}{!}{\begin{tabular}{@{}llcccccrrrr@{}}
\toprule
Model & Labels & Exact & Bal. & F1 & G/P & $\pm1$ & \multicolumn{4}{c}{Recall P/CL/CY/O} \\
\midrule
Historical (2025) & 6,147 hist. & 63.6 & 59.2 & 56.7 & 85.6 & 93.2 & 77.0 & 36.2 & 31.1 & 92.6 \\
A: same labels, new recipe & 6,147 hist. & 62.7 $\pm$ 0.6 & 58.2 $\pm$ 0.6 & 55.2 $\pm$ 1.1 & 85.2 $\pm$ 0.5 & 92.7 $\pm$ 0.6 & 76.5 & 36.1 & 26.2 & 93.9 \\
B: size-matched crowd & 6,147 crowd & 66.2 $\pm$ 0.8 & 61.8 $\pm$ 1.0 & 60.2 $\pm$ 1.1 & 88.2 $\pm$ 0.5 & 95.0 $\pm$ 0.6 & 75.5 & 44.6 & 32.2 & 95.0 \\
C: full pool & 14,049 crowd & 70.2 $\pm$ 0.5 & 67.5 $\pm$ 0.3 & 66.9 $\pm$ 0.4 & 89.5 $\pm$ 0.5 & 96.6 $\pm$ 0.2 & 75.0 & 52.1 & 58.8 & 84.0 \\
D: full pool, YOLO11n & 14,049 crowd & 71.0 $\pm$ 0.2 & 68.0 $\pm$ 0.3 & 67.7 $\pm$ 0.2 & 89.7 $\pm$ 0.2 & 96.8 $\pm$ 0.1 & 72.9 & 55.0 & 59.3 & 85.0 \\
\bottomrule
\end{tabular}
}
\end{table}

\paragraph{Model renewal.}
The crowd labels were consolidated into a four-class pool of 14{,}049 frames (11{,}934 training, 2{,}115 validation; \cls{Bad} excluded; majority label, a single vote for 96\%) and a frozen held-out set of 3{,}720 readable frames from 33 cameras. The split assigns whole seven-day blocks, shared by all cameras, to training, validation or test; test blocks contain no frame of any earlier training set, adjacent blocks can fall in different splits, and pool frames within an hour of a same-camera held-out frame are dropped (median gap to the nearest same-camera training frame five days); the evaluation is within-camera and within-period. \Cref{tab:downstream} separates the three things that changed between the original classifier, a YOLOv8n~\cite{jocher2023yolov8} trained by the original study~\cite{bible2026fujiview} on the 6{,}147-frame split recovered with its checkpoint, and its renewal: the recipe, the labels and their number. Rows A--C share one YOLOv8n recipe (ImageNet initialisation, $224\times224$ input, AdamW, early stopping within 100 epochs); D uses YOLO11n~\cite{jocher2024yolo11}. Retraining on the historical labels under this recipe lands within a point of the original checkpoint (A: $62.7\pm0.6$ versus 63.6), so the recipe does not account for the subsequent improvement. Replacing them by the same number of crowd labels with the same training class histogram (B) adds 3.4 points of exact accuracy (Welch~\cite{welch1947generalization} 95\% interval $[2.4,4.5]$) and 5.0 of macro-F1; growing the crowd set to the full pool (C) adds another 4.1 ($[3.1,5.1]$) and 6.8. Over the original checkpoint the renewed regime gains 6.6 points of exact accuracy and 10.3 of macro-F1, mostly where the labels are hardest (\cls{Clear} and \cls{Cloudy} recall). The gains associated with renewed supervision and additional labels are much larger than the YOLOv8n-to-YOLO11n backbone gain: D improves on C by only 0.7 points (Welch 95\% interval $[0.2,1.3]$ over five seeds each).

\emph{Interpretation and scope.} A and B differ in images, labels and validation split (271 shared; B's validation histogram deviates by 76 frames; A early-stops on its original random split), and same-camera historical frames lie within 24~h of 41\% of the held-out frames, which favours A without helping it (B leads A by 4.5 points on them and by 2.7 beyond a week), so A$\rightarrow$B measures renewed labels on a renewed sample, not label quality alone. The held-out targets are single crowd votes, so A, trained on a single annotator's labels, is also penalised where that annotator's boundaries differ from the crowd's; the gain survives on the coarse \cls{Good}/\cls{Poor} decision (+3.0, $[2.3,3.7]$), where the two agree about 90\%, and the ceiling of \cref{sec:noise} applies to every row. And held-out exact accuracy is not comparable with the 47--58\% human agreement on gold frames, whose class mix differs (41\% of held-out frames are \cls{Obscured}, the most consistently labeled class).

\paragraph{Label-budget curve.}
We retrained the D recipe (YOLO11n) on nested 5\%, 10\%, 25\% and 50\% subsets of the pool, five times. Each replicate fixes one random ordering of the training frames and one of the validation frames, of which every budget takes the first fraction (unstratified, nested), trains with its own seed and is scored on the same held-out frames; its 100\% point is the D run with that seed (\cref{fig:budget}, right). With 703 labels (5\% of the pool) the recipe reaches $61.2\pm2.0\%$ exact accuracy (mean $\pm$ SD over replicates, range 58.5--63.6); 1{,}405 labels give $65.0\pm0.9\%$, 3{,}513 labels $68.2\pm0.4\%$, 7{,}025 labels $70.2\pm0.3\%$ and the full 14{,}049 labels $71.0\pm0.2\%$. The curve is concave in log label count and stable across replicates: every step raised exact accuracy in every replicate but one (a 0.05-point dip); the first two steps ($\times2$, $\times2.5$) bought 3.8 and 3.2 points, the last two doublings $2.0\pm0.6$ and $0.8\pm0.3$. Relative to its own full-pool run, a quarter of the pool delivers on average 96.1\% of the full-pool exact accuracy (95\% interval $[95.2,96.9]$; at least 94.9\% in every replicate; 2.8 points below it, or 91\% of the gain over the 40.8\% majority class) and 95.0\% of its macro-F1, and half the pool 98.9\%. In effort, the campaign's 23{,}894 labels took about 55 active labeler-hours (30-minute sessionisation), so the 25\% point corresponds pro rata (1.7 campaign labels per pool label) to roughly 14 labeler-hours and the 5\% point to about 3. Replicated budget curves can thus inform annotation stopping points, especially where small budgets have more variable outcomes.

\section{Limitations and Conclusion}
\label{sec:limitations}
\label{sec:conclusion}

The sessions were an in-house campaign, not a randomised experiment, and although the wall was available to the remote session the tool did not log who viewed it, so the gamification results are descriptive; the participants were hourly-paid students at one institution, and the engagement patterns need not generalise to other populations or incentives. The reference set is one author's labels of frames from three days in May 2025, so agreement with it measures consistency, not correctness, and its exact level may be specific to those conditions. The held-out set uses single crowd votes and its test weeks are interleaved with training weeks of the same cameras (63\% of its frames have a same-camera training frame within a week), so we claim within-camera, within-period model renewal, not generalisation to unseen cameras or later periods; and we tested no stronger pretrained encoder, so the label-versus-backbone conclusion holds for this backbone family only. 

The gamified annotation system showed encouraging signs of sustained participation: in the closest remote comparison, labelers remained active longer with similar throughput and reference agreement, although causation cannot be inferred. Label disagreement concentrated near ordinal boundaries, while within-one-class and \cls{Good}/\cls{Poor} decisions were more stable, including for the reference annotator. The largest model gains accompanied renewed supervision and a larger annotation pool, while the backbone swap yielded a smaller gain. A quarter of the pool achieved 96\% of full-pool exact accuracy on average. We plan to release scripts, manifests, and non-identifying derived results following privacy and sharing review.


\bibliographystyle{splncs04}
\bibliography{references}

\end{document}